\documentclass[10pt]{article}
\usepackage[english]{babel} 
\usepackage{subcaption} 

\usepackage{authblk}

\usepackage{iftex}
\ifPDFTeX
    \usepackage[utf8]{inputenc}
\else
\fi

\usepackage[margin=1.1in]{geometry}
\usepackage[T1]{fontenc} 
\usepackage{amsmath, amssymb} 
\usepackage{graphicx}
\graphicspath{{Figures/}} 
\usepackage[colorlinks=true, allcolors=blue]{hyperref} 
\usepackage{csquotes} 
\usepackage[ backend=biber, style=nature, citestyle=numeric-comp, sorting=none ]{biblatex} 
\begin{document}

\title{\LARGE A Saccade-inspired Approach to Image Classification using Vision Transformer Attention Maps}

\author[1,2,*]{Matthis DALLAIN}
\author[2]{Laurent RODRIGUEZ}
\author[1]{Laurent Udo PERRINET}
\author[2]{Beno\^it MIRAMOND}

\affil[1]{Institut de Neurosciences de la Timone, Aix-Marseille Universit\'e, CNRS, Marseille, 13005, France}
\affil[2]{Laboratoire d'\'Electronique, Antennes et T\'el\'ecommunications, Universit\'e C\^ote d'Azur, CNRS, Sophia Antipolis, 06903, France}

\affil[*]{corresponding author: matthis.dallain@univ-cotedazur.fr}

%\keywords{Vision Transformer, Neuro-inspired AI, Artificial Saccades, Image Classification, Energy-efficient Computer Vision}
\maketitle

\begin{abstract}

Human vision achieves remarkable perceptual performance while operating under strict metabolic constraints. A key ingredient is the selective attention mechanism, driven by rapid saccadic eye movements that constantly reposition the high resolution fovea onto task-relevant locations, unlike conventional AI systems that process entire images with equal emphasis. Our work aims to draw inspiration from the human visual system to create smarter, more efficient image processing models. Using DINO, a self-supervised Vision Transformer that produces attention maps strikingly similar to human gaze patterns, we explore a saccade-inspired method to focus the processing of information to key regions in visual space. 
To do so, we use the ImageNet dataset in a standard classification task and measure how each successive saccade affects the model’s class scores. This selective-processing strategy preserves most of the full-image classification performance and can even outperform it in certain cases. By benchmarking against established saliency models built for human gaze prediction, we demonstrate that DINO provides superior fixation guidance for selecting informative regions. These findings highlight Vision Transformer attention as a promising basis for biologically inspired active vision and open new directions for efficient, neuromorphic visual processing.
\end{abstract}

\flushbottom

\section*{Introduction}

To cope with the amount of visual information it is confronted with at any given moment, the primate visual system relies on the small but high-resolution central region of the retina known as the \emph{fovea}, where a high density of cone photoreceptors yield greatest visual acuity \autocite{yan_cell_2020}. Visual signals from this region travels to the cortex, where it is allocated over‑represented cortical resources, a phenomenon known as cortical magnification \autocite{virsu_cortical_1987, daniel_representation_1961, chaplin_representation_2013, wassle_cortical_1989, frisen_optical_1975}. It is believed that this mechanism ensures that computational and metabolic cost of visual processing is devoted mainly to this foveal input. Rapid eye movements, or saccades, then shift the fovea across a scene, enabling detailed sequential perception while minimizing energy use \autocite{yarbus_eye_2013, land_oculomotor_2011,schutz_eye_2011, gegenfurtner_interaction_2016}. 
In contrast, most artificial vision systems process entire images with uniform resolution, which is computationally and energetically costly, and doesn't take into consideration the redundancy of the visual input. This discrepancy has motivated multiple bio-inspired approaches that mimic saccade-like selective sampling \autocite{larochelle_learning_2010, mnih_recurrent_2014, jonnalagadda_foveater_2022, lukanov_biologically_2021, dauce_dual_2020, rasamuel_specialized_2019} to reduce computational cost while maintaining competitive performance.

This link between biological and artificial attention is particularly relevant in the context of Vision Transformers (ViTs) \autocite{dosovitskiy_image_2021}, whose core mechanism, self-attention, computes weighted interactions between image regions (referred to as \enquote{image patches}), and selectively extract task relevant information, akin to how humans prioritize informative areas of a scene. Specifically, the information contained in each patch is weighted, extracted and fused into a high dimensional vector, drawing a pattern of information flow (often referred to as an \enquote{attention map}) that highlights parts of the image that the model identified as important for the task. 

Recently, it has been shown that the DINO (Distillation with NO labels) self-supervised ViT \autocite{caron_emerging_2021} produces interpretable attention maps that align to a significant degree with human gaze \autocite{yamamoto_emergence_2025, carrasco_vision_2025, djilali_vision_nodate}. Without explicit supervision or ground-truth labels, DINO learns to focus on semantically meaningful regions of images, such as faces or other contextually important elements, indicating that semantically grounded attention can emerge naturally through unsupervised learning. 

Building on this idea, we implement a sequential, attention-based sampling strategy in which image regions are progressively revealed based on their attention scores. Using the attention patterns from a pretrained DINO Vision Transformer, we select the most informative regions at each step and feed them to a classifier. By measuring classification performance as a function of the number and locations of revealed regions, we evaluate whether attention-driven saccades capture meaningful and discriminative content. This approach allows us to test the hypothesis that sparse, ViT's attention-guided visual sampling can achieve efficient recognition, analogously to the energy-efficient sampling strategy of the human visual system.

Here, we do not set out to build or train a fully operational model, nor do we aim for state-of-‑the-‑art accuracy. Rather, we concentrate on exploring the intrinsic power of attention patterns —their capacity to pinpoint informative regions for classification, their general characteristics, and how they can be harnessed to reduce computational load. We also benchmark these patterns against saliency maps produced by models explicitly built for human‑gaze prediction, using them as a reference point. This work is intended as an accessible entry point to Vision Transformer's attention maps for readers interested in the synergy between biological and artificial vision. By showcasing their interpretability and functional usefulness, we aim to demonstrate how such maps can serve as a practical, low‑cost tool for vision research.

\section*{Related work}

\subsection*{Models of visual saliency}

Theories of visual attention suggest that gaze allocation results from the integration of bottom-up, saliency-driven cues and top-down, goal- or semantics-driven influences \autocite{itti_computational_2001}. Early computational models mainly addressed the bottom-up component, predicting fixations from low-level image features such as color, contrast, or motion. The seminal model by Itti, Koch, and Niebur \autocite{itti_model_1998} and its successors, such as the Graph-Based Visual Saliency (GBVS) model \autocite{harel_graph-based_2006}, compute feature maps from basic visual dimensions, normalize them, and integrate them into a saliency map. These models capture stimulus-driven guidance effectively but lack mechanisms to account for task or contextual influences on gaze.

With the advent of deep learning, saliency prediction evolved toward data-driven approaches that learn to associate visual content with recorded human fixation patterns \autocite{yan_review_2022, borji_saliency_2019}. Models such as DeepGazeII \autocite{kummerer_deepgaze_2016} and SALICON \autocite{huang_salicon_2015} learn hierarchical visual representations that implicitly encode semantic and contextual information \autocite{yan_review_2022, kummerer_deepgaze_2022}. Although their architectures remain \enquote{bottom-up} in structure, they achieve a closer match to human gaze behavior by reflecting low-level saliency as well as high-level regularities in the data \autocite{hayes_deep_2021}. 

Interestingly, the self-supervised DINO model have been shown to produce attention maps that align closely with human gaze patterns \autocite{yamamoto_emergence_2025, carrasco_vision_2025, djilali_vision_nodate}, even though they are trained without any supervision from eye-tracking data. DINO’s attention exhibits both semantic selectivity as well as bottom up characteristics, such as odd-one-out detection in artificial displays, that are absent in deep saliency models \autocite{yamamoto_emergence_2025}, suggesting a spontaneous emergence of saliency-like mechanisms within general visual representations \autocite{djilali_vision_nodate}.

\subsection*{Attention maps for efficient image processing} 

Vision Transformers process non-overlapping image patches, transformed into independent vector (or patch tokens), without relying on spatial adjacency. This position-agnostic design makes them more robust to occlusion \autocite{naseer_intriguing_2021} and enables selective processing of most informative tokens \autocite{nguyen_token_2025}.
Interestingly, several studies have shown that ViTs attention map naturally highlights the most relevant parts of an image. For instance, Naseer et al. \autocite{naseer_intriguing_2021} reported that classification accuracy remains nearly unchanged even when up to 50\% of the less attended tokens are masked.

These findings have motivated a series of token-pruning strategies that exploit the model's attention map to concentrate computation on salient regions. Evo-ViT \autocite{xu_evo-vit_2021}, for example, removes background tokens from the attention graph and updates them using representative tokens that remain connected to the global context. Liang et al. \autocite{liang_not_2022} proposed an inattentive token fusion mechanism, which merges the least-attended tokens into an attention-weighted average while preserving the top-$k$ informative ones. Other approaches have explored attention-guided dynamic resolution. Chen et al. \autocite{chen_cf-vit_2022} observed that ViTs maintain stable attention patterns even under reduced image resolutions and proposed an early-exit mechanism: if classification fails at a coarse resolution, high-attention patches are subdivided for finer analysis and a second pass through the model is realized. Building on this principle, Hu et al. \autocite{hu_lf-vit_2024} proposed to select entire regions rather than isolated patches, for detailed reprocessing, similarly to the high resolution fovea–low-resolution periphery structure of human retina. This hierarchical scheme further reduces token redundancy while maintaining classification performance.

\subsection*{Existing saccade models}

Multiple models have explored processing images through a sequence of fixations, or \enquote{glimpse}, to reduce computational cost and mimic biological saccades. In contrast to the soft-attention mechanism used in Vision Transformers (ViTs), which assigns continuous weights to all parts of the input, these models employ a hard-attention strategy, selecting only a subset of the input while discarding completely the rest \autocite{guo_attention_2022}. Because this selection process is non-differentiable, various strategies have been proposed to train models to efficiently decide where to \enquote{glimpse}.

Early work by Schmidhuber et al. \autocite{schmidhuber_learning_1990} demonstrated that fixation control and recognition modules could be jointly trained through backpropagation in a visual search task. Later, Larochelle and Hinton \autocite{larochelle_learning_2010} used a Restricted Boltzmann Machine trained via Contrastive Divergence \autocite{carreira-perpinan_contrastive_nodate} on more complex benchmarks such as MNIST.
A popular strategy introduced by Butko and Movellan  \autocite{butko_infomax_2010} employed policy gradients to learn optimal sequences of glimpses, paving the way for deep learning approaches such as the Recurrent Attention Model, by Mnih et al \autocite{mnih_recurrent_2014}. Although these models initially struggled to scale to natural images \autocite{sermanet_attention_2015}, recent advances have shown that saccadic strategies can be effective even on challenging datasets such as ImageNet \autocite{elsayed_saccader_2019,jeremie_integrating_2025, huang_glance_2022}, and have proven suitable for fine grained classification task, for images that are visually and semantically close to one another \autocite{liu_fully_2017, zhao_survey_2017}.

In contrast to these methods, ViTs rely on a differentiable attention mechanism where attention maps naturally emerge from the optimization objective: routing task-relevant information across tokens. Close to our approach, the Foveater model \autocite{jonnalagadda_foveater_2022} leveraged these attention maps to guide a multi-resolution input strategy, mimicking the human retina by defining high-resolution central tokens and low-resolution peripheral ones. While effective, Foveater recomputes the attention map at each step. Here, we explore whether a single-pass, static saliency map can drive all subsequent saccades, reducing computational overhead. Furthermore, Foveater reported that low resolution peripheral features contributed most to accuracy, making saccade selection efficiency hard to evaluate. This is further accentuated by the fact that attention maps can be noisy depending on the model setup \autocite{darcet_vision_2024, jiang_vision_2025}, a caveat that the DINO model avoids.

\section*{Method}

\subsection*{Vision Transformer}

\begin{figure}[ht]

    \includegraphics[width=\linewidth]{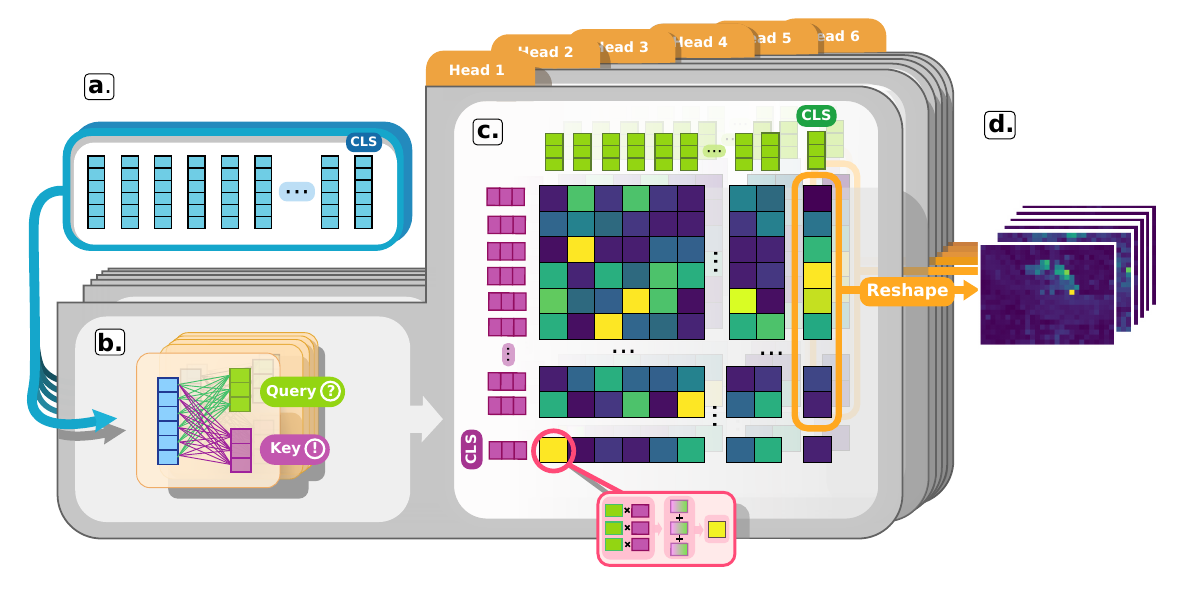}% , height=220

    \caption{Schematic illustration of attention map extraction within a single transformer layer. (\textbf{a.}) The input set of token embeddings, including the additional \texttt{[CLS]} token, is passed trough the multiple attention heads for parallel processing. (\textbf{b.}) Within each head, the tokens are projected through learned linear layers to obtain Query, Key, and Value representations (Value projections omitted in the figure for clarity). (\textbf{c.}) Attention weights are computed as the scaled dot product between Queries and Keys (as depicted in the bottom pink frame). The last column of the resulting attention matrix corresponds to the \texttt{[CLS]} token, and yields a vector of attention scores indicating the relative importance assigned to each image patch. (\textbf{d.}) This vector is reshaped into a 2-D grid to recover the spatial layout of the original image. This operation outputs one spatial attention map per head.}
    \label{fig:Attention Map Extraction}

\end{figure}

\begin{figure}[ht]

    \includegraphics[width=\linewidth]{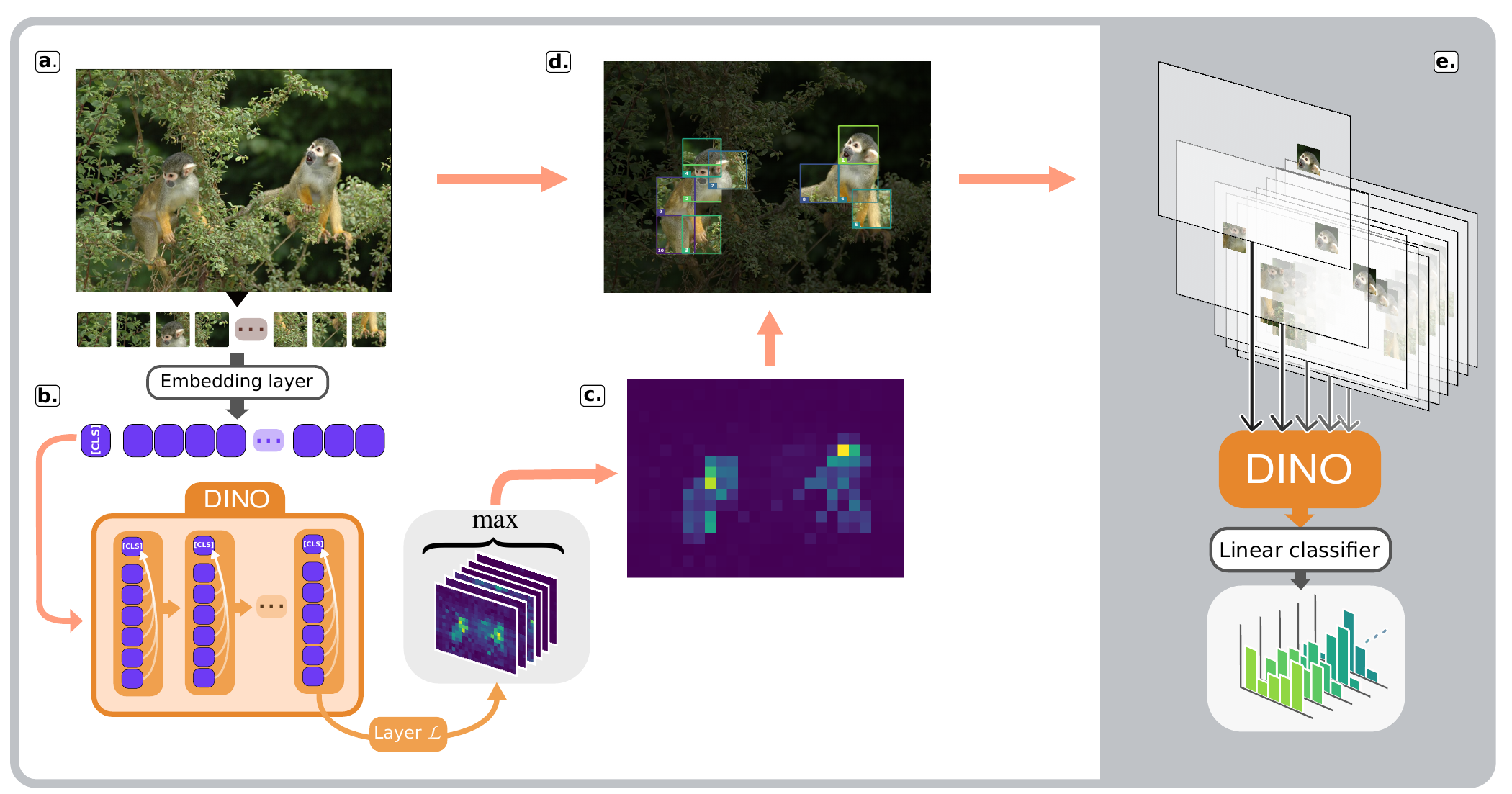}% , height=220

    \caption{Saccade selection method: (\textbf{a.})~The input image of dimension \(H \times W\) is split into {\tiny $\dfrac{H}{16} \times \dfrac{W}{n}$} sized patches and embedded into token vectors. (\textbf{b.})~The tokens are passed through the DINO transformer, and attention flow from patch tokens to \texttt{[CLS]} token (white arrows) are extracted and reshaped into one attention map per attention-head. (\textbf{c.})~The multiple attention maps are fused into one by taking the maximum value across heads. (\textbf{d.})~The highest-attention locations define square regions (\enquote{saccades}) whose tokens are retained. (\textbf{e.})~Selected regions are revealed sequentially, and the image variants are classified by a pre-trained linear head. Image adapted from Wikimedia Commons \cite{gdml_saïmiri}}
    \label{fig:method}

\end{figure}

Vision transformers (ViTs) process images by first dividing it into a set of N non-overlapping patches of $n \times n$ pixels (Figure~\ref{fig:method}.a). These are then projected into a high dimensional space via a series of learnable linear filters, resulting in $N\times D_{\text{emb}}$ vectors, referred to as patch tokens. A learned additional vector (the \texttt{[CLS]} token), unrelated to the image, is also added to the initial set, which purpose is to accumulate task-relevant information from every other patch tokens (Figure~\ref{fig:method}.b). At this stage, each token only encodes local visual information, as no interaction exists between patches. Additionally, the positional structure of the image is lost through this patchification process, and needs to be reintroduced by adding a $N\times D_{\text{emb}}$ spatial embedding tensor to the original set of tokens, thus biasing their initial values according to their position on the image. 

The ensemble thus forms a $(N+1) \times D_{\text{emb}}$ matrix $X$ to be processed by the $L$ layers of the model. Each layer is composed of the Self Attention (SA) module (illustrated in Figure~\ref{fig:Attention Map Extraction}), and the subsequent feed-forward network $FFN$, composed of two linear layers with a non-linearity in between. Within the SA module, the matrix $X$ is processed in parallel by the so called \enquote{attention heads}, where each of those $n_h$ heads applies its own set of learned linear projections to the full input $X$, producing three set of lower-dimensional vectors, called Queries, Keys, and Values (Figure~\ref{fig:Attention Map Extraction}.b):

\[
Q_i = X W_Q^{(i)}, \quad K_i = X W_K^{(i)}, \quad V_i = X W_V^{(i)},
\]

where $W_Q^{(i)}, W_K^{(i)}, W_V^{(i)} \in \mathbb{R}^{D_{\text{emb}} \times D_h}$, with $D_h = D_{\text{emb}} / n_h$, are the learned weight matrices specific to the $i^{th}$ attention head. The query, key, and value vectors can be interpreted as a question, a measure of token relevance to that question, and the information to be transmitted in response to that question, respectively. Specifically, for a given token, the similarity between its query vector and the key vectors of all tokens determines how much attention the former should pay to these latter. Indeed, these similarity scores are used to weight the corresponding value vectors, which contain the information to be aggregated and propagated across tokens. This weighted flow of information is what define the attention mechanism, allowing each patch token to be contextualized based on the content of all other tokens. It is formally defined by:
\[\text{Attention}(Q, K, V) = \text{softmax}\left( \frac{QK^\top}{\sqrt{D_h}} \right) V.\]

where the dot product between the Queries and Keys term produces a matrix of vector similarity (Figure~\ref{fig:Attention Map Extraction}.c), or \enquote{attention weight matrix}, for each given head $i$:
\[
A_i = \text{softmax} \left( \frac{Q_i K_i^\top}{\sqrt{D_h}} \right) \in \mathbb{R}^{(N+1) \times (N+1)}.
\]

The similarity matrix is transformed by a softmax operation, turning raw scores into positive weights that sum to 1. This not only normalizes the values but also introduces competition between tokens, since increasing the attention toward one patch necessarily decreases attention toward others. 
Each row of $A_i$ thus contains the attention weights from one token to all others. In particular, the row corresponding to the query \texttt{[CLS]} token (e.g. the first one) encodes how much attention it gives to each patch token to achieve its task related role.

The self-attention map of the \texttt{[CLS]} token for head $i$ is defined as:

\[
\mathbf{a}_i^{\texttt{[CLS]}} := \left[ A_i \right]_{1,\,2:N+1} \in \mathbb{R}^N,
\]

where we take the first row of $A_i$ and exclude the first column (which would be the token attending to itself). This yields a 1D attention map over the $N$ patch tokens (The $A_i$ matrix is modified for clarity in Figure~\ref{fig:Attention Map Extraction}.d).

This vector can be reshaped to match the original spatial arrangement of the image patches (e.g., into a $\sqrt{N} \times \sqrt{N}$ grid), providing a spatially interpretable \emph{attention map} that shows where the \texttt{[CLS]} token \enquote{looks} in the image (Figure~\ref{fig:Attention Map Extraction}.d).

Finally, the weighted $D_{\text{emb}} / n_{h}$ Value vectors from all \( n_h \) heads are reshaped into the original embedding dimension $D_{\text{emb}}$ and projected via another learned linear transformation, before being added to the previous patch tokens through a skip connection, and being processed by the $FFN$. This attention-weighted combination of Values thus allows each token to incorporate contextual information across the entire image based on the relation it shares with the others. 

This process repeats for each of the $L$ layers, and at the end the \texttt{[CLS]} token is used as a compact, weighted representation of the input image, which is then passed through a projection head for classification.

\subsection*{Dataset}

We use the ImageNet-1K dataset \autocite{deng_imagenet_2009} validation set for our classification task. Following the pipeline defined by the DINO paper, we first resize the images so that its smaller dimension is set to 256 pixels, preserving the aspect ratio, after which we apply a \(224 \times 224\) pixels center crop. The images are then normalized using the ImageNet mean and standard deviation values.

\subsection*{Saccade mechanism}

We first pass the original, unaltered image through the DINO model to obtain the resulting attention map from a given transformer layer \(L\) (typically the last layer, \(L = 12\), unless otherwise stated) (Figure~\ref{fig:method}.b). Since the input images have a resolution of \(224 \times 224\) pixels, and the patch are of size \(16 \times 16\) pixels, the resulting attention maps are of size \(\mathbb{R}^{14 \times 14}\). To reduce the multi-head attention maps into a single map, we take the maximum value across heads at each spatial location (Figure~\ref{fig:method}.c).
The location of the maximum attention score within the \(\mathbb{R}^{14 \times 14}\) grid is then used to define a square region of fixed size (referred to as a \enquote{fovea} here), either \(3 \times 3\) or \(5 \times 5\) tokens (regions of \(48 \times 48\) or \(80 \times 80\) pixels, respectively) in the original image. These regions are extracted from the already tokenized representation of the image and centered on the identified coordinates (Figure~\ref{fig:method}.d). 
After a region is sampled, the corresponding area in the attention map is suppressed by setting its values to a negative constant. This prevents subsequent saccades from repeatedly selecting the same location, mimicking the human visual system’s inhibition-of-return mechanism \autocite{itti_computational_2001}.

\subsection*{Computational Considerations}

Generating attention maps requires passing the image through DINO twice, once to compute the attention map and once for classification. This strategy is used in this paper to allow the study of self-attention-based saccades, but is not computationally efficient. We therefore explore alternative, less demanding settings. First, we extract attention maps from earlier layers to test whether saccades can be guided by cheaper, less refined features. Second, we evaluate lower-resolution inputs (\(224 \times 224\) pixel inputs resized to \(128 \times 128\), \(112 \times 112\), or \(96 \times 96\) pixels), which yield attention maps of \(8 \times 8\), \(7 \times 7\), and \(6 \times 6\) tokens, respectively. The smaller attention maps are then resized back to a \(14 \times 14\) grid through bilinear interpolation. For these experiments, we'll use only the \(3 \times 3\) token fovea when selecting the saccades.

\subsection*{Classification}

The cropped regions, selected through the previously defined saccade mechanism, are then used as successive inputs for an image classification task. For each image, we thus generate ten variants corresponding to sequential \enquote{saccades}, each containing an additional region of interest to the ones already selected, before passing them through the model and eventually to a pretrained linear classifier head (Figure~\ref{fig:method}.e). This allows us to evaluate how informative the selected saccades are for object recognition in a given image. 

We follow the classification pipeline originally defined for the DINO model: for each image, we extract the \texttt{[CLS]} token from the 4 last layers, concatenate them and feed the result into a linear classifier. The classifier and model weights are directly loaded from the official DINO GitHub repository \autocite{noauthor_facebookresearchdino_nodate}.

We additionally compute the relationship between the certainty of the model's predictions and the number of saccades necessary to achieve correct classification, and compare it between random and attention-driven fixations. For each classification vector \(p\) of n values (here, n=1000), we compute the certainty \(C(p\)) as the 1-normalized entropy, that is, equal to one for a maximal entropy, and zero for a minimal value:
\[
C(p) = 1 - \frac{\sum_{i=1}^{n} p_i \log(p_i)}{\log(n)}
\]

\subsection*{Control Experiments}

As a control, we reproduce the selection procedure using a randomly generated map of values, such that regions are sampled at random locations (\enquote{random fixations}) rather than guided by attention.
To assess whether accuracy improvements arise from uncovering informative regions (guided by attention) or simply from the progressive increase in visible tokens, we record the percentage of visible pixels for each image variant. 

We also consider the possibility that images correctly classified after only a few saccades exhibit low-entropy attention maps, where high attention values are concentrated within a small region corresponding to the object of interest. In such cases, a few fixations may be sufficient to capture the most discriminative information, as most of the attention flow from patch tokens to the \texttt{[CLS]} token originates from this localized area. This will help us determine whether saccades mainly serve to uncover an object of interest or actually find informative regions that allows it to extrapolate beyond its current view.

Additionally, verifying this relationship between attention-map entropy and classification efficiency could inform a more adaptive sampling strategy, where the number of saccades allocated to the model is determined by the entropy of the attention distribution which would reduce the need for recurrent sampling when the model’s focus is already well-defined.

To test this assumption, we compute the entropy of each attention map. For an attention map flattened to a distribution \(p\), the entropy is defined as
\[
H(p) = - \sum_{i}^{n} p_i \log (p_i).
\]

\section*{Results}

\subsection*{Classification Accuracy Across Saccades}

\begin{figure}[ht]

    \centering
    \makebox[\textwidth][c]{%
        \begin{subfigure}{\textwidth}
            \includegraphics[width=\linewidth]{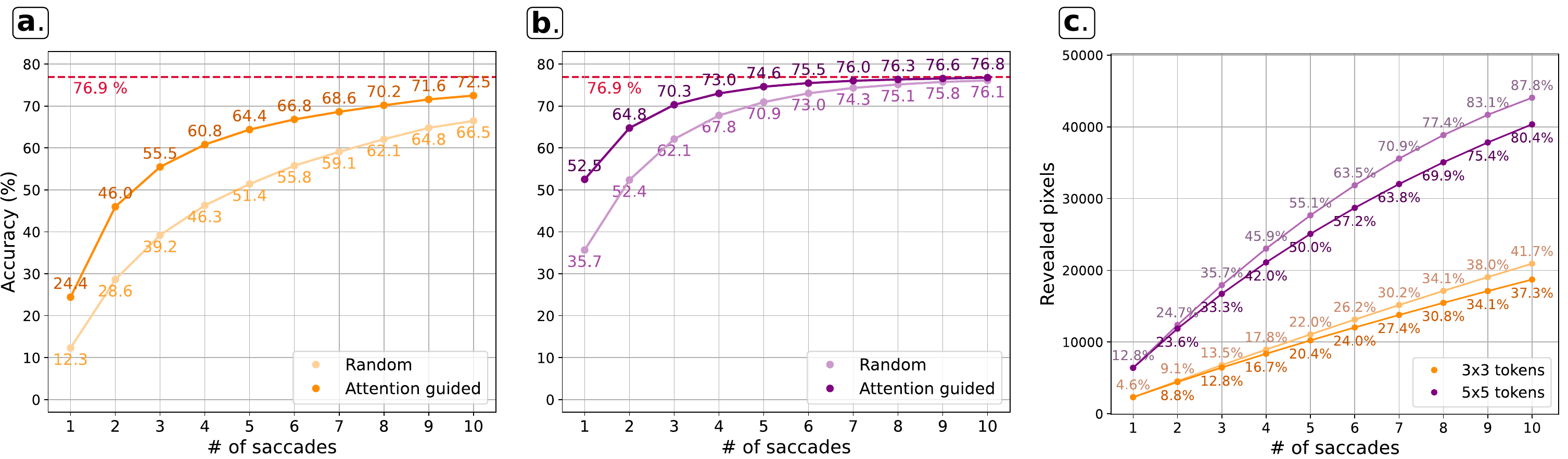}
        \end{subfigure}
        
    }
     
    \caption{Accuracy score across saccades: For a fovea of size 3x3 tokens (\textbf{a.}), 5x5 tokens (\textbf{b.}), and the corresponding percentage of the image given to the model throughout saccades (\textbf{c.}). The accuracy score when the model is given the full image is shown as a dotted red line. The light curves correspond to the random saccades in every plot.}
    \label{fig:Accuracy 3x3 5x5}
\end{figure}

Several consistent patterns emerge when examining classification accuracy. As more image regions are revealed through successive saccades, accuracy steadily increases across the dataset (Figure~\ref{fig:Accuracy 3x3 5x5} a. and b.). This trend appears in both attention-guided and random sampling, but the increase is markedly steeper in the attention-guided condition—especially during the first few saccades—indicating that attention maps effectively target informative regions. Beyond a certain number of fixations, however, performance plateaus and the gap between attention-guided and random sampling narrows, suggesting diminishing returns once sufficient image content has been explored. Random fixations eventually approach the performance of attention-guided ones, but only after revealing substantially more tokens.

Larger fovea sizes accelerate accuracy gains, reaching near-original performance after only a few saccades (Figure~\ref{fig:Accuracy 3x3 5x5}.b), however accuracy improvements cannot be explained solely by the number of tokens revealed. Indeed, attention-guided fixations typically expose fewer unique tokens than random ones (Figure~\ref{fig:Accuracy 3x3 5x5}.c)— due to more frequent spatial overlap between successive regions—yet still achieve higher accuracy, reflecting that the highest attention values align at least to some extent with the most informative areas. 

\subsection*{Classification Dynamics}

\begin{figure}[ht]

    \makebox[\textwidth][c]{%
        \includegraphics[width=1.05\textwidth]{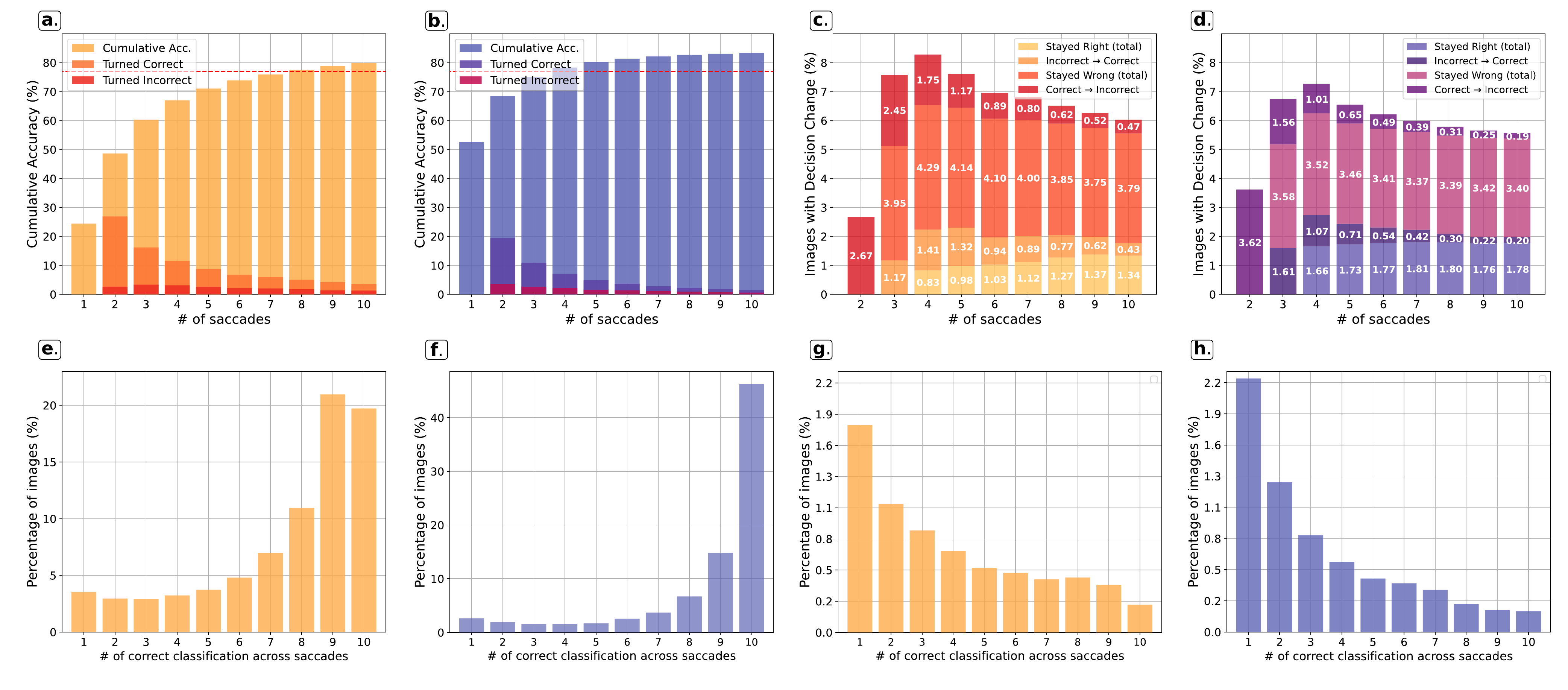}
    }

    \caption{Histogram of classification dynamics across saccades: \textbf{a.}  and \textbf{b.} Cumulative accuracy (percentage of images correctly classified at least once), with the proportion of images that turned correct or incorrect at each saccade. \textbf{c.}  and \textbf{d.}. Percentages of images that changed classification at this specific saccade (stacked bars) (Correct$\rightarrow{}$Wrong and Wrong$\rightarrow{}$Correct) and the proportions that have remained incorrect (Stayed Wrong) or remained recovered to correct (Stayed Right). \textbf{e.}  and \textbf{f.} Percentage of images distributed according to the number of occurrence of correct classification across saccades, over the full validation dataset, and over the subset of images that were correctly classified during saccades but not when the image was given full (\textbf{g.}  and \textbf{h.}). The percentage correspond to the percentage over the whole validation dataset in every case. \textbf{a.}, \textbf{c.}, \textbf{e.}  and \textbf{g.} : \(3 \times 3\) tokens fovea. \textbf{b.}, \textbf{d.}, \textbf{f.}  and \textbf{h.}: \(5 \times{} 5\) tokens fovea}
    \label{fig:Cumulative Acc.}

\end{figure}

An unexpected finding emerged when evaluating classification at the image level rather than at each individual saccade: When considering the cumulative accuracy ---the progressive count of images that were correctly classified at least once across saccades---, the overall classification score actually surpasses that obtained when the model is given the full image (Figure~\ref{fig:Cumulative Acc.}, a. and b.), an effect even more pronounced in the case of the \(5 \times 5\) fovea (Figure~\ref{fig:Cumulative Acc.}, b.). 

To understand this, we examined the dynamics of the model’s decision changes across saccades. Specifically, at each saccade, we identified:
\begin{itemize}
    \item the set of images that switched from correct to incorrect (Correct$\rightarrow{}$Wrong)
    \item the set of images that, after a first change, switched back from incorrect to correct (Wrong$\rightarrow{}$Correct)
    \item the set of images that remained incorrect after switching from correct in previous saccades (Stayed Wrong)
    \item the set of images that remained correct after switching from incorrect in previous saccades (Stayed Right)
\end{itemize}

We observe that most classification changes occur during the first few saccades (Figure~\ref{fig:Cumulative Acc.}, c. and d.), although the model continues to revise its decisions throughout the sequence. A substantial proportion of images that initially became incorrectly classified are later restored to a correct label, indicating some capacity for recovery (Figure~\ref{fig:Cumulative Acc.}, c. and d.). However, the size of the Stayed Wrong and Stayed Wright sets tends to remain stable, even if we observe that, over time, the size of the former slightly decreases while the later gradually increases (although see latest saccades in Figure~\ref{fig:Cumulative Acc.}, d.). 

These observations may partly stem from noise, as while most images are consistently classified correctly across the saccade sequence (Figure~\ref{fig:Cumulative Acc.}, e. and f.), images correctly identified only during saccades were generally recognized only a few times (Figure~\ref{fig:Cumulative Acc.}, g. and h.). A non negligible percentage still appears correctly classified for the majority of saccades, however.

\subsection*{Certainty Evolution Across Saccades}

\begin{figure}[ht]
    \centering
    \makebox[\textwidth][c]{%
        \begin{subfigure}{0.9\textwidth}
            \includegraphics[width=\linewidth]{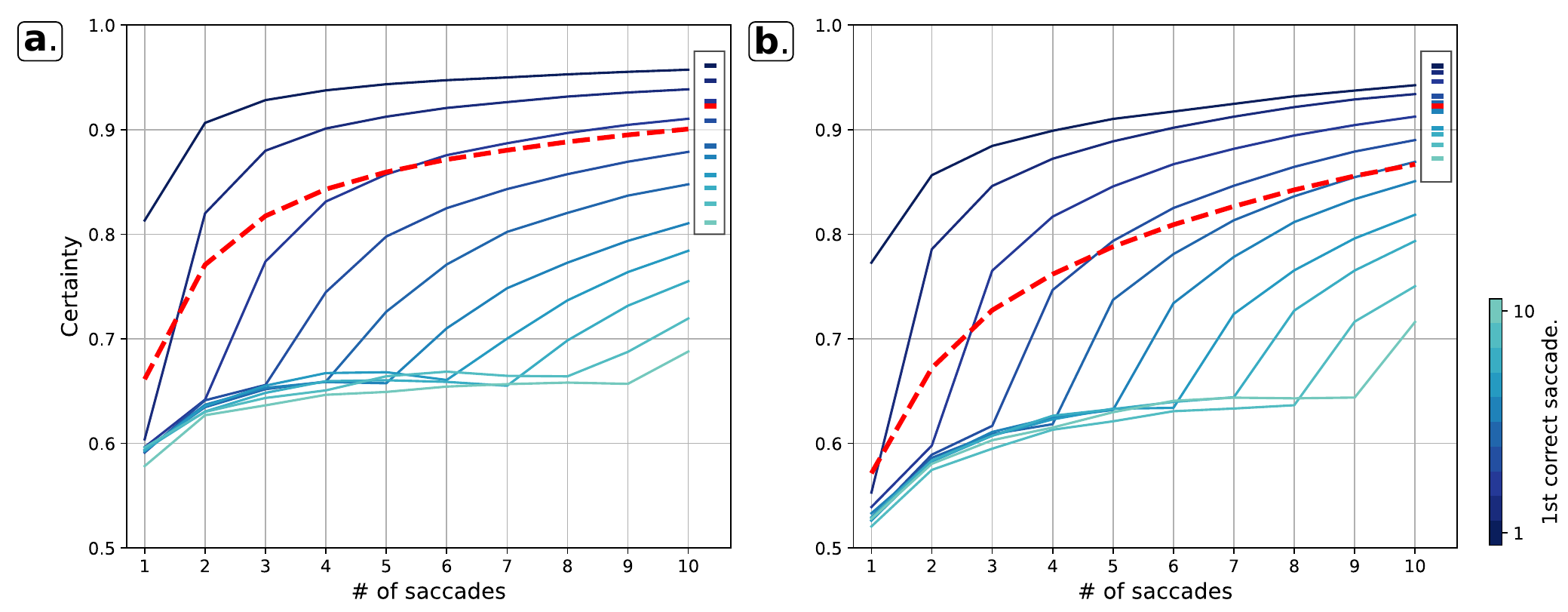}

        \end{subfigure}
    }
     
    \caption{Evolution of the model’s classification certainty across saccades: Curves represent the mean certainty for attention driven (\textbf{a.}) and random saccades (\textbf{b.}). Colors indicate sets of images correctly classified at a given saccade (dark to light blue: first to tenth saccade). The red dotted line shows the mean certainty of all images, across saccade. Colored bars in the rectangles on the right of each plot shows mean certainty when the full image is given to the model, for each set.}
    \label{fig:Certainty}
\end{figure}

To assess confidence across saccades, we restricted our analysis to images that the model correctly classified under full-input conditions, ensuring a fair comparison between full image and saccade classifications, and measured the entropy of the output distribution at each saccade. We found that the relationship between attention and informativeness is further supported by certainty measures: mean certainty across saccades (Red dotted curves in Figure~\ref{fig:Certainty}) remains higher under attention-driven sampling (Figure~\ref{fig:Certainty} a.) and closely matches that obtained when the full image is available (red bar in the rectangles in Figure~\ref{fig:Certainty} a. and b.). In contrast, random fixations yield lower certainty even though the images were correctly classified, showing that although the model can classify accurately from random crops, it tends to do so less confidently (Figure~\ref{fig:Certainty} b.).

We then divided the images into different sets, based on the saccade at which first correct classification occurred (blue curves in Figure~\ref{fig:Certainty}), while images that were never correctly identified during saccades were discarded. Again, we measured the certainty at each saccade, for every set, and although images are grouped by the saccade of their first correct classification, certainty was still evaluated for all subsequent saccades. Additionally, we measured the certainty when the entire images were given to the model, for each set (blue bars in the rectangles in Figure~\ref{fig:Certainty}). We repeat this analysis for random and attention driven cases.

%\vspace{3mm}

In both conditions, mean certainty shows a sharp increase at the moment of correct classification for the different sets, but continues to rise across successive saccades, and eventually plateaus near the level reached when the full image is visible --or below for the sets correctly classified at later saccades (light blue curves)-- (Figure~\ref{fig:Certainty} a. and b. colored bars), showing that correct classification doesn't necessarily occur when the model is most confident. It also shows that, in average, adding more information comfort the model in his decision.

Finally, we observe that images correctly identified after only a few saccades tend to be those the model classifies with highest certainty when viewing the full image (darkest blue bars in Figure~\ref{fig:Certainty}), suggesting they are inherently easier examples for the model.

\subsection*{Layer-wise and Resolution Analyses}

We observed that attention maps become increasingly effective at guiding saccades as we progress through the layers of the model (Figure~\ref{fig:Accuracy Across Layers, Across Resolutions}, a.). The first layer actually performs worse than for random saccades, while the second one produces accuracy above chance. We observe a sharp increase in accuracy after the third layer, reaching performance levels closer to those obtained when using the final layer, which indicate that low-level features already contain some spatially informative cues. The number of revealed pixels remains comparable across layers (Figure~\ref{fig:Accuracy Across Layers, Across Resolutions}, b.), although the effect of layer depth becomes more pronounced as saccades accumulate —likely because deeper attention maps are more refined and less noisy than earlier ones (see Appendix Figure~\ref{figS:Examples attmaps through layers}).

As the resolution of the attention maps decreases, overall performance drops slightly, yet attention-guided saccades remain markedly more effective than random fixations (Figure~\ref{fig:Accuracy Across Layers, Across Resolutions}, c.). Notably, with \(8 \times 8\) tokens attention maps, almost half as many tokens, accuracy remains close to that obtained in the full-resolution setting. We see that the amount of revealed pixels tends to decrease as the resolution of the attention map is reduced (Figure~\ref{fig:Accuracy Across Layers, Across Resolutions}, d.), an effect which may partly stem from their rescaling to a \(14 \times 14\) grid used for saccade selection. Indeed, upscaling produces smoother transition between the peaks of attention values, high-attention peaks spreading over a larger neighborhood. The usual \(3 \times 3\) suppression therefore may not be enough to cancel their influence, so multiple saccades tend to fall within the same area instead of exploring new parts of the image.

The combined effect of depth and resolution can be seen in the Appendix Figure~\ref{figS:Accuracy across layers low res}. We observe similar trends, although with worse performance overall. Still we see that less refined features of low resolution input are also capable of identifying informative regions in an image.
 
 \begin{figure}[ht]
    \centering
    \makebox[\textwidth][c]{%
        \begin{subfigure}{1.05\textwidth}
            \includegraphics[width=\linewidth]{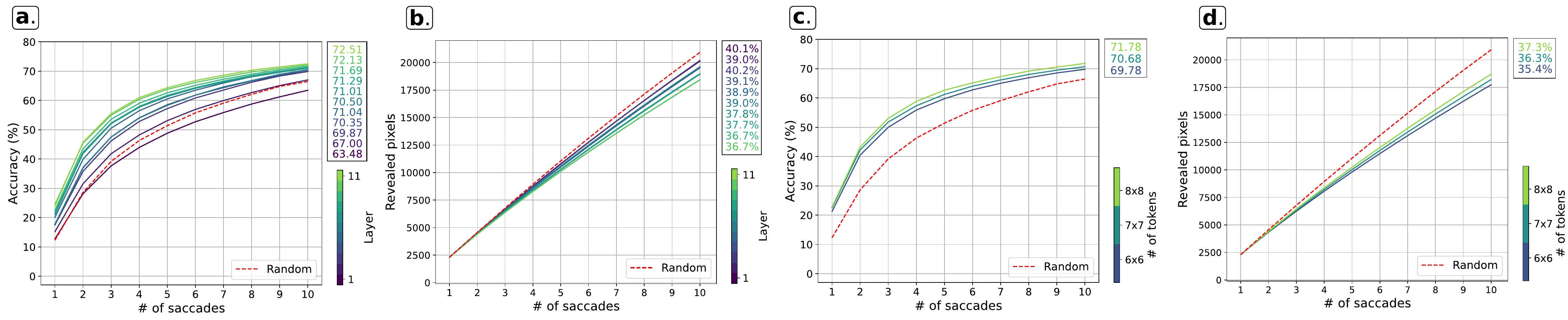}
        \end{subfigure}
    }
     
    \caption{Saccade accuracies for attention maps across layers and input resolutions: Accuracy scores for attention maps from different layers (\textbf{a}) and the corresponding percentage of the image revealed during saccades (\textbf{b}). Accuracy scores for attention maps from different input resolutions (\textbf{c}) and their corresponding revealed pixels (\textbf{d}). Red dotted lines indicate results for random fixations.}
    \label{fig:Accuracy Across Layers, Across Resolutions}
\end{figure}

\subsection*{Entropy of attention maps}

\begin{figure}[ht]
    \centering
    \makebox[\textwidth][c]{%
        \begin{subfigure}{0.9\textwidth}
            \includegraphics[width=\linewidth]{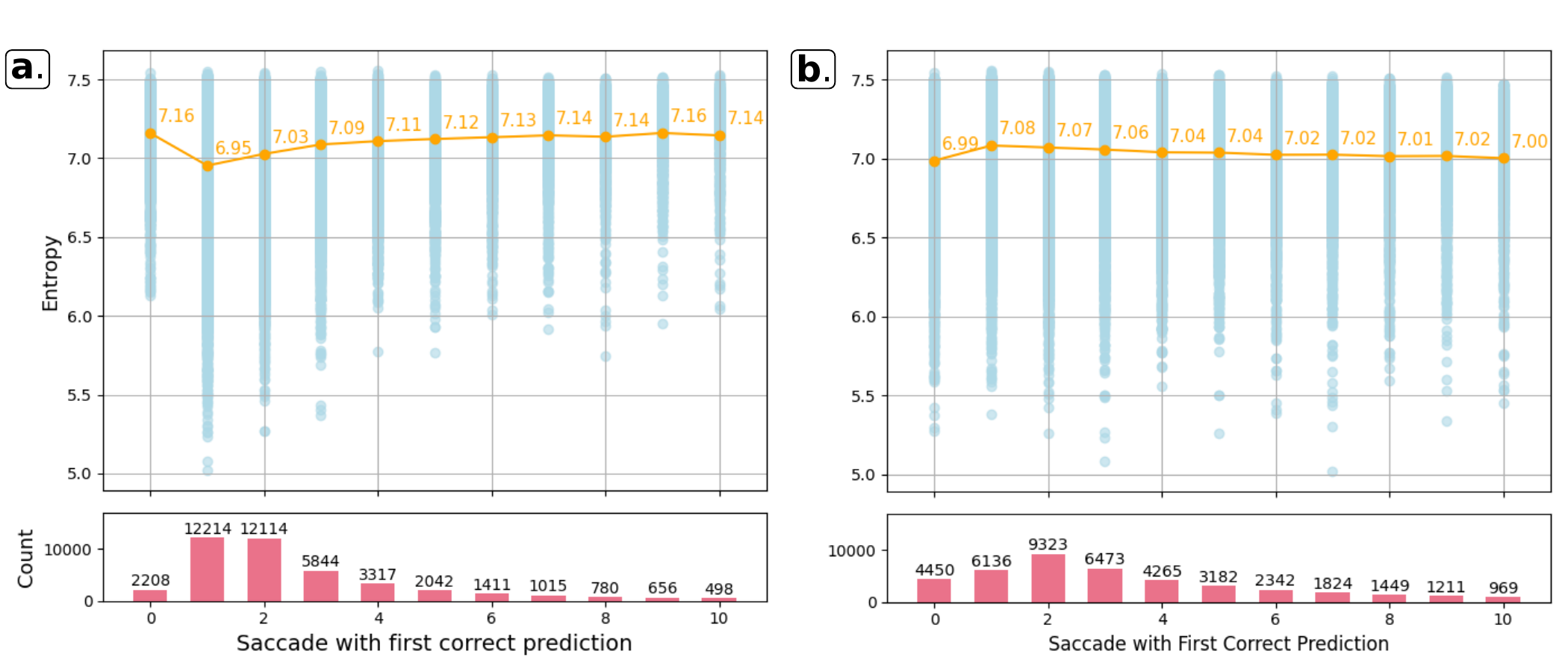}
        \end{subfigure}
    }
     
    \caption{Entropies of attention maps grouped by the saccade leading to a correct classification : The first column (Saccade 0) shows images correctly classified when given full to the model, but not during saccades. The orange line indicates the mean entropy across saccades. The lower panel shows the number of images processed at each saccade. (\textbf{a.} attention-driven fixations. \textbf{b.} randomly selected fixations.)}
    \label{fig:Entropy per saccade}
\end{figure}

Analysis of the attention maps reveals a slight positive relationship between entropy and the number of saccades required for correct classification in the attention-guided condition, whereas the opposite trend is observed for random saccades, as we initially hypothesized (Figure~\ref{fig:Entropy per saccade}). Moreover, among images that remain unclassified, none in the attention-driven condition exhibit entropy values below 6.0, while such low-entropy maps are present in the random-fixation condition, showing that random saccades might be more likely to miss an object of interest in a given image than attention driven saccade.

\begin{figure}[ht]

    \includegraphics[width=\linewidth]{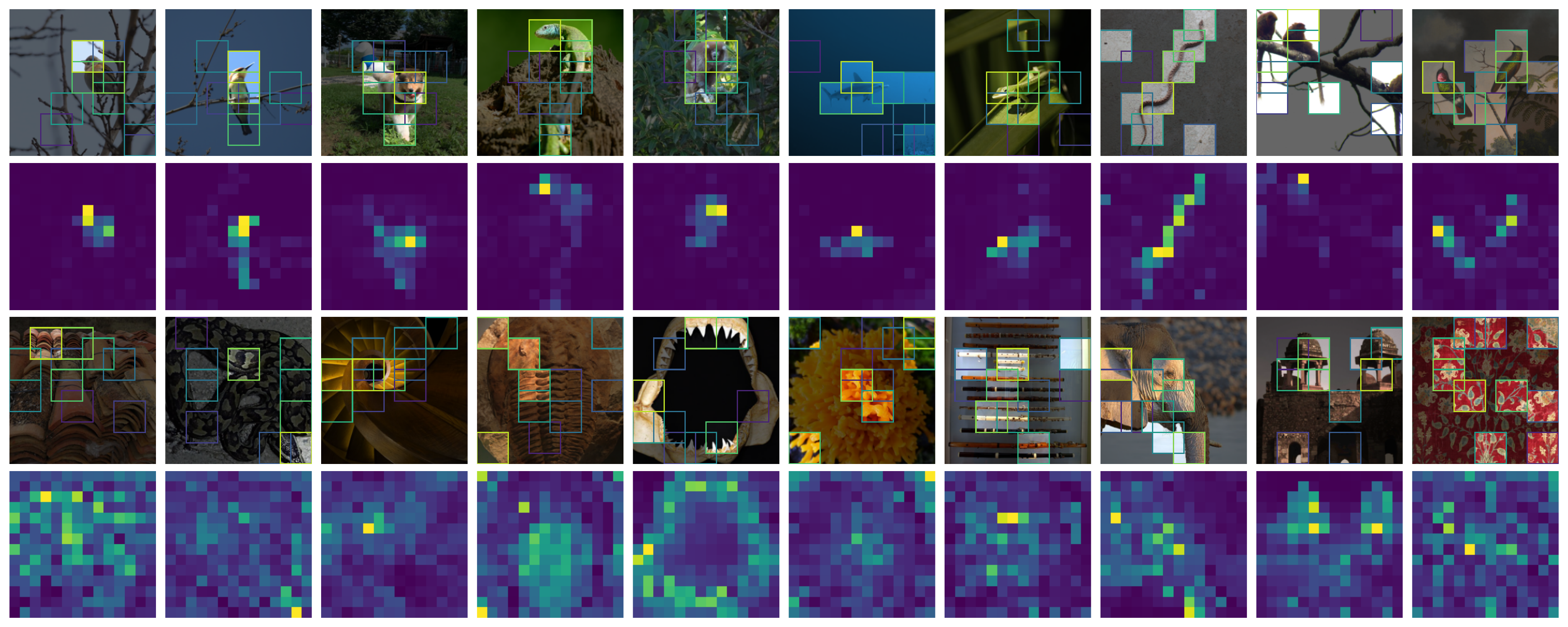}

    \caption{Examples of images correctly classified at different numbers of saccades, with their corresponding attention maps: Upper and lower rows display low- and high-entropy attention maps, respectively. Saccade order is coded as a colored square from light green (first) to dark blue (tenth); fixations leading to correct classification are highlighted. Images adapted from Wikimedia Commons \cite{states_green, pakistan_brambling, yercaudelango_chestnut, liasmith93_a, skot_ještěrka, johnmartindavies_elphinstone, area_hypsiglena, robertson_brown, heade_martin, kawakami_plecturocebus, milošević_wooden, bel_tejas, graham_coral, vitvit_čmh, tschentscher_female, india_unknown, anupamg_purana, koenig_bonn, robertson_carcharodon, todd_fossil}}
    \label{fig:Saccade pattern per entropy}

\end{figure}

Although low-entropy maps generally concentrate attention on compact and coherent regions, they do not always ensure rapid classification. Even when the object of interest falls entirely within a saccade, the model sometimes requires additional fixations to reach the correct decision—possibly because contextual cues surrounding the object are needed for disambiguation, or because cropping perturbs the classification process (Figure~\ref{fig:Saccade pattern per entropy}). Conversely, some high-entropy maps still lead to correct classification within only a few saccades, indicating that the model can still infer the correct label from incomplete information.

Thus, despite establishing a link between the number of saccade before correct classification and attention map entropy, those results contradict our initial intuition that entropy could reliably predict how many tokens are required for accurate classification.

\subsection*{Comparison with Visual Saliency Prediction Models}

To determine whether the regions identified as informative by DINO’s attention maps are model-specific or contain features that are generally informative for image recognition, we performed two complementary analyses.

First, we evaluated whether these regions also support accurate classification by a different architecture. Specifically, we used a pretrained ResNet-50 model \autocite{he_deep_2015} and cropped the \(48 \times 48\) pixel region (2.1\% of the image) selected by DINO’s attention maps. Each cropped region was then classified independently, and top-1 accuracy was recorded. Second, we replaced DINO’s attention maps with saliency maps generated by external models to guide the saccade selection process, in our original sequential saccade pipeline as well as the ResNet-50 classifier one. For comparison, we also tested two simple baselines in which the cropped region was selected either at the center of the image or at a randomly chosen location.

These experiments allowed us to assess whether the regions identified by DINO are uniquely tailored to its own internal representations, or whether alternative biologically-inspired saliency maps can substitute for the model’s own attention in guiding the progressive reveal of image regions.

We considered two types of saliency models:
\begin{itemize}
    \item The Graph Based Saliency Model (GBVS) \autocite{scholkopf_graph-based_2007-1} as a representative classical bottom-up model. We used a python implementation made available on GitHub \autocite{shreenath_shreelockgbvs_2025}.

    \item The Unified Image and Video Saliency Modeling (UNISAL) \autocite{droste_unified_2020, droste_rdrosteunisal_2025}. model, as a modern deep saliency model. It has been shown to better predict human gaze than DINO's attention maps \autocite{djilali_vision_nodate} and as such provides a useful baseline for assessing whether more human-like saccadic behavior would enhance classification.
\end{itemize}

UNISAL is a versatile model trained to predict gaze position over images as well as videos. As such, it is designed with dataset-specific parameters (e.g. priors, batch normalization, smoothing...). Interestingly, we found that the configuration trained on the DHF1K video dataset \autocite{wang_revisiting_2018} yielded the best results in our setup, and therefore used it for the remainder of the experiments.

Since UNISAL was trained on the DHF1K dataset using images of size \(288 \times 384\) pixels, we resize all inputs to this resolution for both UNISAL and the other models to ensure consistency. The saliency map were all resized through bilinear interpolation, to {\tiny $\dfrac{H}{16} \times \dfrac{W}{16}$} (i.e. \(18 \times 24\)) to match DINO's attention maps.
Due to computational constraints, this set of experiments was conducted on a subset of the ImageNet validation set, created by randomly sampling 10 images from each of the 1,000 classes.

\subsubsection*{Impact on classification accuracy}

\begin{figure}[ht]
    \centering
    \makebox[\textwidth][c]{%
        \begin{subfigure}{1.0\textwidth}
            \includegraphics[width=\linewidth]{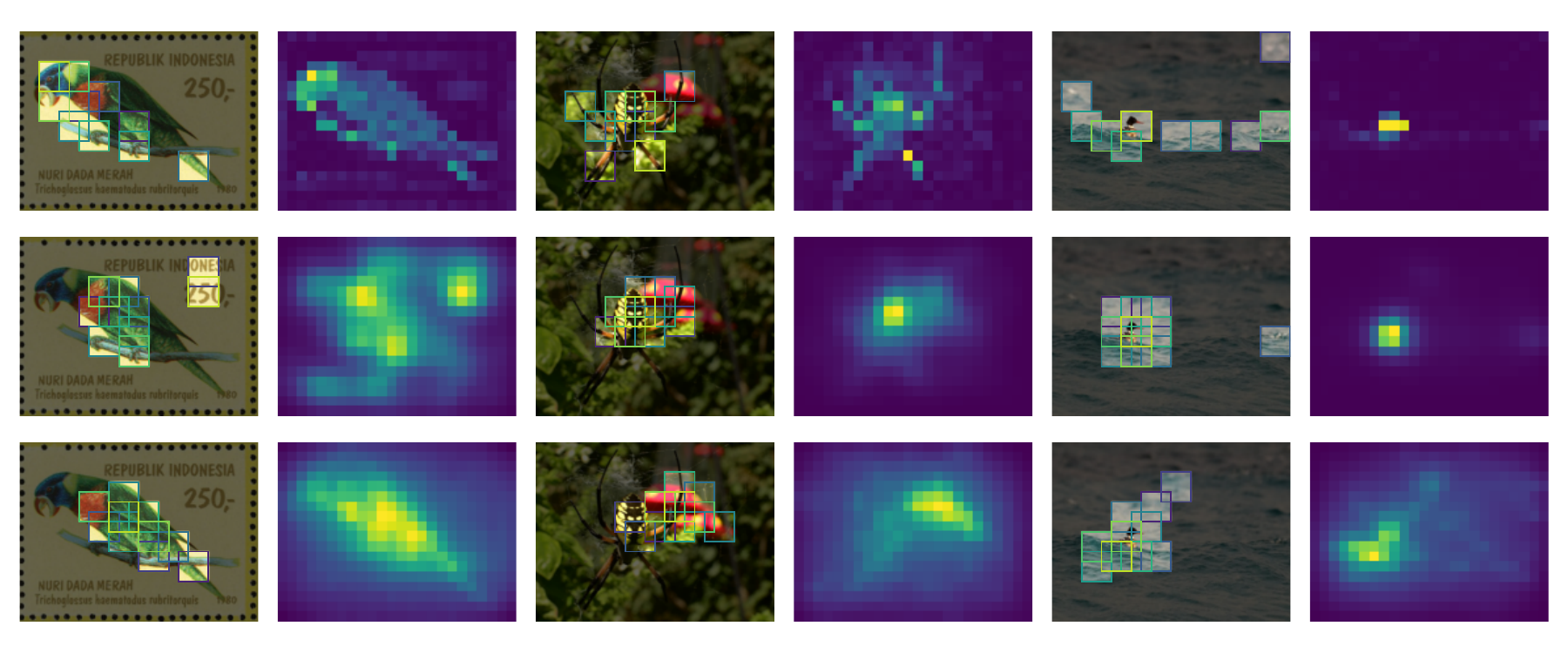}
        \end{subfigure}
        
    }
     
    \caption{Examples of saliency maps generated by the different models and their corresponding saccade patterns. Images adapted from Wikimedia Commons \cite{indonesia_stamp,jeffreyw_argiope,pennington_redbreasted}
    }
    \label{fig:Saliency Different Models}
\end{figure}

\begin{figure}[ht]
    \centering
    \makebox[\textwidth][c]{%
     
        \begin{subfigure}{\textwidth}
            \includegraphics[width=\linewidth]{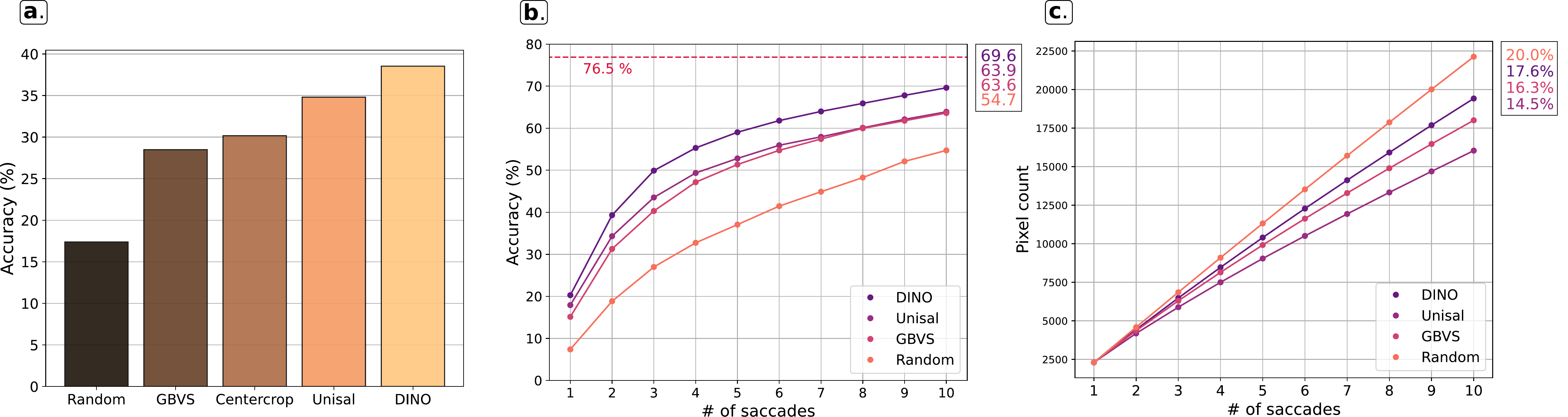}
        \end{subfigure}
        }
    
    \caption{Effect of saliency maps generated by the different models: Accuracy scores of the ResNet50 classifier using saliency maps from different models, as well as random and centered fixations (\textbf{a}). Accuracy scores of the DINO classifier, over multiple saccades driven by these attention maps (\textbf{b}) and the corresponding percentage of the image revealed during saccades (\textbf{c}).
    }
    \label{fig:Saccades Saliency Models}
\end{figure}

Examples of saliency maps obtained from each model are shown in Figure~\ref{fig:Saliency Different Models}. When applied to a CNN classifier, DINO-based attention maps still guide saccades effectively, yielding the highest accuracy among all saliency-driven conditions (Figure~\ref{fig:Saccades Saliency Models}, a.), showing that DINO's attention maps are not model specific, and generalize well across architectures. GBVS saliency also performs better than random crop, however, the performance remains below that of a simple center crop of the image, while both UNISAL and DINO achieve higher accuracy overall.

When considering the saccade selection method, although all saliency models successfully guide fixations to informative areas, DINO consistently maintains around 5 \% higher accuracy than the other two models across all saccades (Figure~\ref{fig:Saccades Saliency Models} b.). GBVS initially performs worse than UNISAL but gradually catches up, reaching comparable accuracy after the sixth saccade.

\begin{figure}[ht]
    \centering
    \makebox[\textwidth][c]{%
        \begin{subfigure}{1.0\textwidth}
            \includegraphics[width=\linewidth]{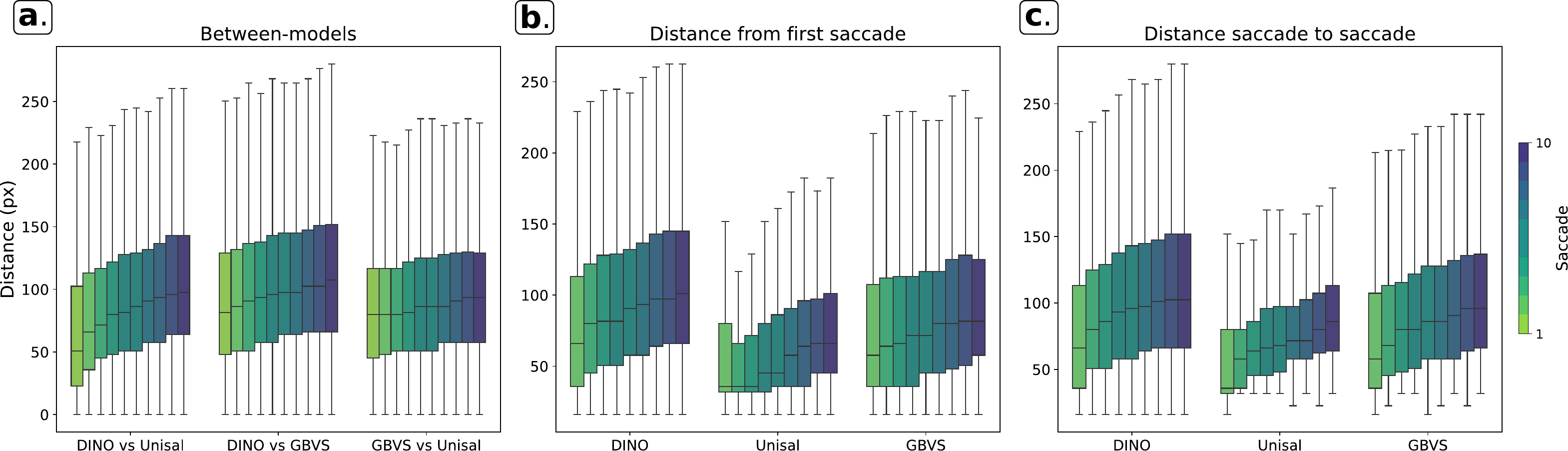}

        \end{subfigure}
        
    }

    \caption{Distances between fixations coordinates across saccades: Between saccades generated by different models (\textbf{a.}), between the first fixation point and the following saccades (\textbf{b.}), and between a given saccade and the preceding one (\textbf{c.}).}
    \label{fig:Distances Saliency Models}
\end{figure}

During the first few saccades, DINO and UNISAL tend to direct fixations toward similar regions, whereas GBVS diverges earlier (Figure~\ref{fig:Distances Saliency Models} a.). As saccades progress, disagreement between all three models increases—a trend particularly pronounced when comparing the two saliency models with DINO.

These differences can be partly explained by the spatial exploration strategies of the models. Both DINO and GBVS reveal a larger portion of the image than UNISAL (Figure~\ref{fig:Saccades Saliency Models} b.) as their saccades tend to move farther from the initial fixation and cover more distant regions. In contrast, UNISAL explores less extensively and remains more spatially confined. (Figure~\ref{fig:Distances Saliency Models}, b. and c.)

Interestingly, while overall accuracy is lower for those larger images than for \(224 \times 224\) ones (see Figure~\ref{fig:Saccades Saliency Models} b. vs Figure~\ref{fig:Accuracy 3x3 5x5} a.), the drop in performance is much smaller for attention-driven saccades than for random ones. This is consistent with the idea that attention mechanisms still efficiently locate informative regions within a vaster frame, whereas the increased image size makes random fixations less likely to fall on such areas.

\section*{Discussion}

The aim of this study was to investigate whether the self-attention mechanism in Vision Transformers (ViTs) \autocite{dosovitskiy_image_2021} could serve as a strategy for selectively process limited regions of an image, in analogy with human saccadic eye movements \autocite{schutz_eye_2011}. To do so, we simulated saccadic sampling guided by the DINO (Distillation with NO labels) \autocite{caron_emerging_2021} self-supervised ViT's attention maps and evaluated how classification accuracy for images from the ImageNet-1K dataset \autocite{deng_imagenet_2009} evolved as progressively larger portions of the image were revealed.

Our results confirmed that ViT attention maps can efficiently identify informative regions of an image. When saccades followed the attention values of these maps, classification scores became more reliable, accuracy increased more rapidly and remained consistently higher across saccades than when they were randomly generated. We found that most of the original accuracy could be recovered using less than half of the image pixels, although full performance was reached only when nearly all visual information was available to the model.

% \vspace{3mm}

Attention proved particularly beneficial for difficult detection cases, such as when target objects were small or embedded within larger images. Indeed, we found that whereas accuracy for random fixations dropped markedly, performance under attention-driven sampling remained stable, which suggests that self-attention mechanisms are especially effective at locating and prioritizing subtle or spatially limited features that might otherwise be overlooked by unguided exploration.

% \vspace{3mm}

We observe that accuracy converges toward - but never fully reaches - the performance obtained when the model is given the full image. Each successive saccade contributes progressively less, reflecting diminishing returns as more of the image is revealed.

However, when evaluating whether an image is correctly classified at least once across all saccades-rather than measuring general accuracy at each given saccade-classification scores actually exceeds the ones obtained with the full image. On its own, this finding is difficult to interpret, but an interesting possibility is that certain image regions are particularly discriminative for specific classes. When the full image is presented, these regions may be diluted by less informative ones, leading the model toward an averaged or ambiguous prediction. In contrast, sequentially revealing the image may allow the model to make correct classifications when only the most informative regions are visible. This relates to the finding of Taesiri et al. \autocite{taesiri_imagenet-hard_2023} that showed that through selective cropping and zooming on an image, a classification accuracy close to 100\% could be achieved on ImageNet, although their focus was on the effect of zooming, and their crop selection was more restrictive. Additionally, the ImageNet dataset contain images with multiple objects, different than the one to classify, which might introduce even more ambiguity when the full image is taken into consideration \autocite{beyer_are_2020, yun_re-labeling_2021}.

% \vspace{3mm}

This points to a potential interaction between soft-attention, which governs information flow between patches, and hard-attention strategies that fully discard parts of the input \autocite{guo_attention_2022}, raising the possibility that selective sampling might not only reduce computational load but also, under some conditions, improve accuracy. However this also shows that the model’s predictions can fluctuate across saccades: an image correctly identified at one step may later be misclassified as additional tokens are incorporated. 

As such, while suppressing portions of the input to improve accuracy is feasible, exploiting this advantage would necessitate the development of an effective early-exit strategy \autocite{bajpai_survey_2025}. Allowing the model to terminate processing once sufficient information has been gathered would naturally lower computational resource consumption. As is, our results serves mainly to illustrate that with a recurrent strategy comes the additional task of knowing when to stop, a challenge that classification in a single pass doesn't have to deal with.

% \vspace{3mm}

Comparisons with other saliency models further highlighted the distinctiveness of DINO’s attention maps, as they consistently outperformed both the classical Graph Based Saliency Model (GBVS) \autocite{scholkopf_graph-based_2007} and the deep saliency model Unified Image and Video Saliency Modeling (UNISAL) \autocite{droste_unified_2020} in guiding informative saccades—even when the DINO backbone was replaced with a ResNet-50 classifier \autocite{he_deep_2015}. This suggests that although DINO’s attention patterns partially overlap with human gaze, this alone does not account for its superior performance. Rather, the distinctive way in which DINO allocates attention appears to be more efficient for classification. In this context, being more “human-like” is not necessarily advantageous; the model’s discriminative precision seems to matter more than its similarity to human viewing behavior.

None of the models were directly optimized for image classification, however, DINO was trained on ImageNet, while UNISAL only used ImageNet for pretraining its stem. It thus remains possible that a deep saliency model trained explicitly to predict human fixations on ImageNet could more closely match DINO’s saliency predictions and achieve similar results. 

Regardless, DINO appears more effective overall in this current setup. However its performance comes at the cost of exploiting a larger portion of the image than UNISAL, a difference that likely stems in part from the different models training task: while UNISAL is trained on human fixation data, which forms smooth spatial distributions of gaze positions \autocite{wang_revisiting_2018}, DINO is trained not to predict gaze but to define a flow of information from the most informative tokens, regardless of their spatial arrangement. As a result, whereas UNISAL generates smoother, single-peaked saliency maps that confine fixations to a few localized areas, DINO's multi-head attention can highlight multiple distinct informative regions simultaneously, allowing a more flexible sampling strategy that might explain its superior guidance of visual exploration.

% \vspace{3mm}

Together, our results indicate that DINO's attention maps not only reflects the spatial distribution of discriminative information but can also serve as a viable mechanism for active, selective visual processing. In this sense, self-attention provides a useful bridge between machine perception and biological vision.

% \vspace{5mm}
\section*{Limitations and Future Directions}

Despite these promising results, several limitations prevent us from claiming that this strategy is applicable to active visual classification.

First, the DINO model was not trained for a saccade-based task, and it remains unclear whether performance could be further improved—or exploration made more efficient—if the model were fine-tuned or trained specifically for sequential visual sampling, or at least match the initial performance.

Our results also suggest that attention-guided saccades are most beneficial during the early fixations, while later ones contribute less. In human vision, foveal focus is continuously informed by lower resolution peripheral context\autocite{stewart_review_2020}.
Incorporating a similar mechanism—such as low-resolution contextual tokens spanning a larger chunk of the image, as in the Foveater approach \autocite{jonnalagadda_foveater_2022}
—could thus deal with that issue, and reduce the number of fixations required for confident predictions.

Another limitation concerns how classification is recomputed independently at each fixation, which is both biologically implausible and computationally inefficient. Future implementations could integrate recurrence, for example by retaining previously computed keys and values across all layer-a strategy known as KV-cache \autocite{pope_efficiently_2022}-or by introducing recurrent states (e.g., GRUs \autocite{cho_properties_2014} or LSTMs \autocite{hochreiter_long_1997}) to accumulate information over time. Such architectures would also enable adaptive stopping rules based on confidence growth, akin to decision processes in human vision.

The most critical limitation, however, lies in the current two-pass pipeline: one forward pass is required to compute attention maps, and another for classification, limiting scalability. Yet, our analyses show that attention maps from early layers or low-resolution inputs already guide saccades effectively. Leveraging these representations to select informative regions before classification could yield a far more efficient hierarchical scheme—one that performs classification only once on the selected tokens, drastically reducing computation without sacrificing accuracy.

Addressing these limitations would bring this framework closer to a biologically inspired and computationally efficient model of active vision, opening promising directions for future work.

\section*{Author contributions statement}

M.D. designed and implemented the computational framework and analyses, prepared the figures, and drafted the manuscript. L.R., L.P and B.M. proposed the initial analysis that motivated the study, provided conceptual input, recommended additional analyses, and contributed to manuscript revisions. All authors reviewed the final manuscript.

\section*{Funding}

This work is supported by a public grant overseen by the French National Research Agency (ANR) as part of the \enquote{PEPR IA France 2030} program (Emergences project ANR-23-PEIA-0002).

\section*{Data availability}

This study uses the ImageNet-1K dataset, which is accessible for research purposes at the official website \url{https://image-net.org/}(registration required). In addition, a small dataset of images was assembled by adapting images obtained from Wikimedia Commons. All reused images are licensed under Creative Commons CC BY or CC BY-SA and are fully attributed in the bibliography; Image adaptations were limited to standard preprocessing and illustrative overlays (e.g., cropping, resizing, transparency, and region highlighting), as described in the Methods section. The illustrative image dataset was intended for illustrative purposes only, and is therefore not publicly released. No additional experimental datasets were generated.

\section*{Competing interests}

 The author declare no competing interests.

%\printbibliography

\clearpage
\appendix
\section{Appendix}

\begin{figure}[ht]

    \includegraphics[width=\linewidth]{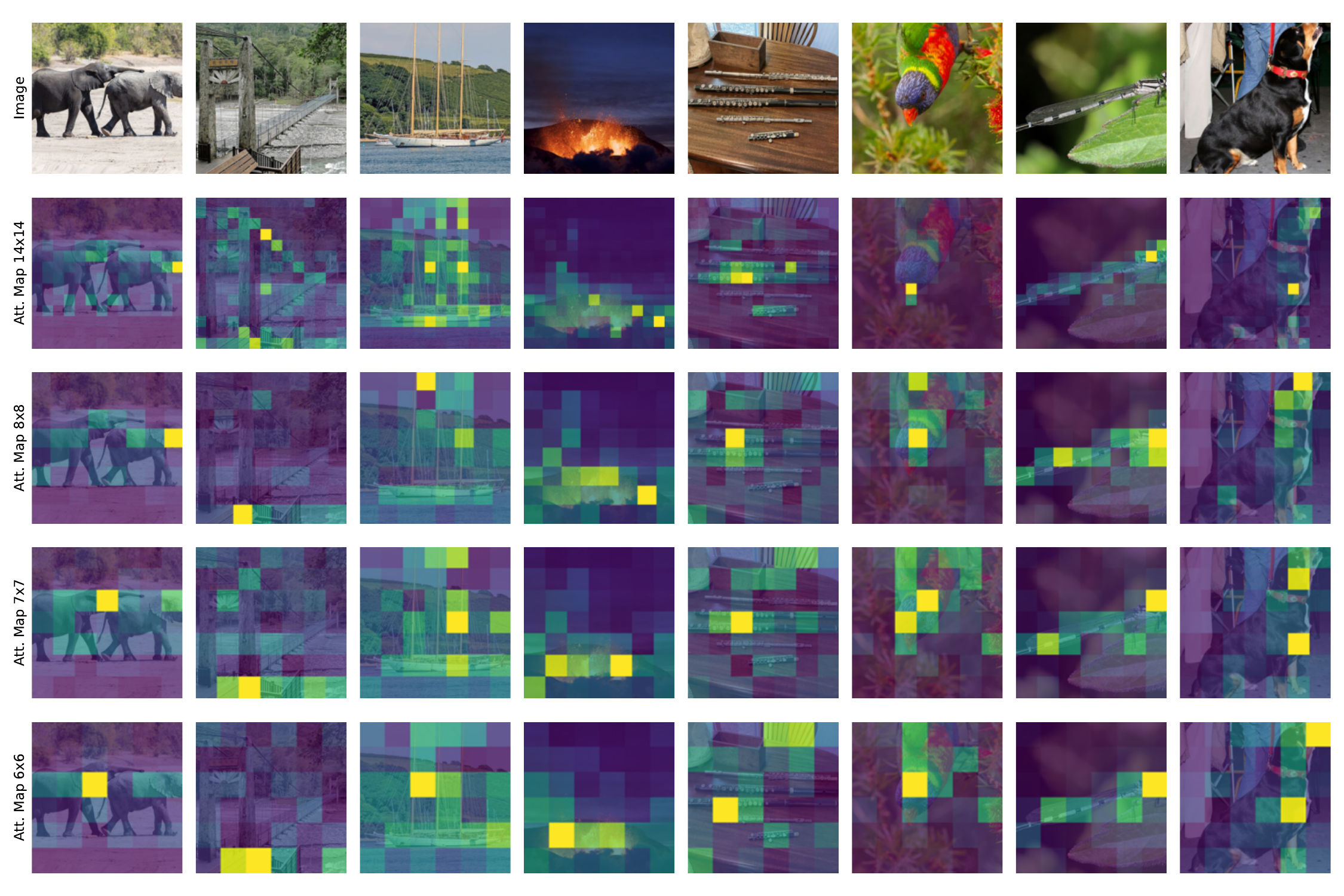} % , height=220

    \caption{Examples of the effect of input resolution on the attention patterns. This figure illustrate the attention maps ability to identify informative regions even from low-resolution input. Once obtained, the low-resolution attention maps are resized to a $14\times{}14$ grid through bilinear interpolation, and are no longer as coarse. All images are from Wikimedia Commons under CC BY or CC BY-SA licenses \cite{delso_elefantes,green_schooner,sharp_common,gagnon_forest,baldwhiteguyconz_rainbow,boaworm_fimmvorduhals,pleple2000_entlebucher,aviv_antique}.}
    \label{figS:Examples attmaps through resolutions}

\end{figure}

\begin{figure}[ht]

    \includegraphics[width=\linewidth]{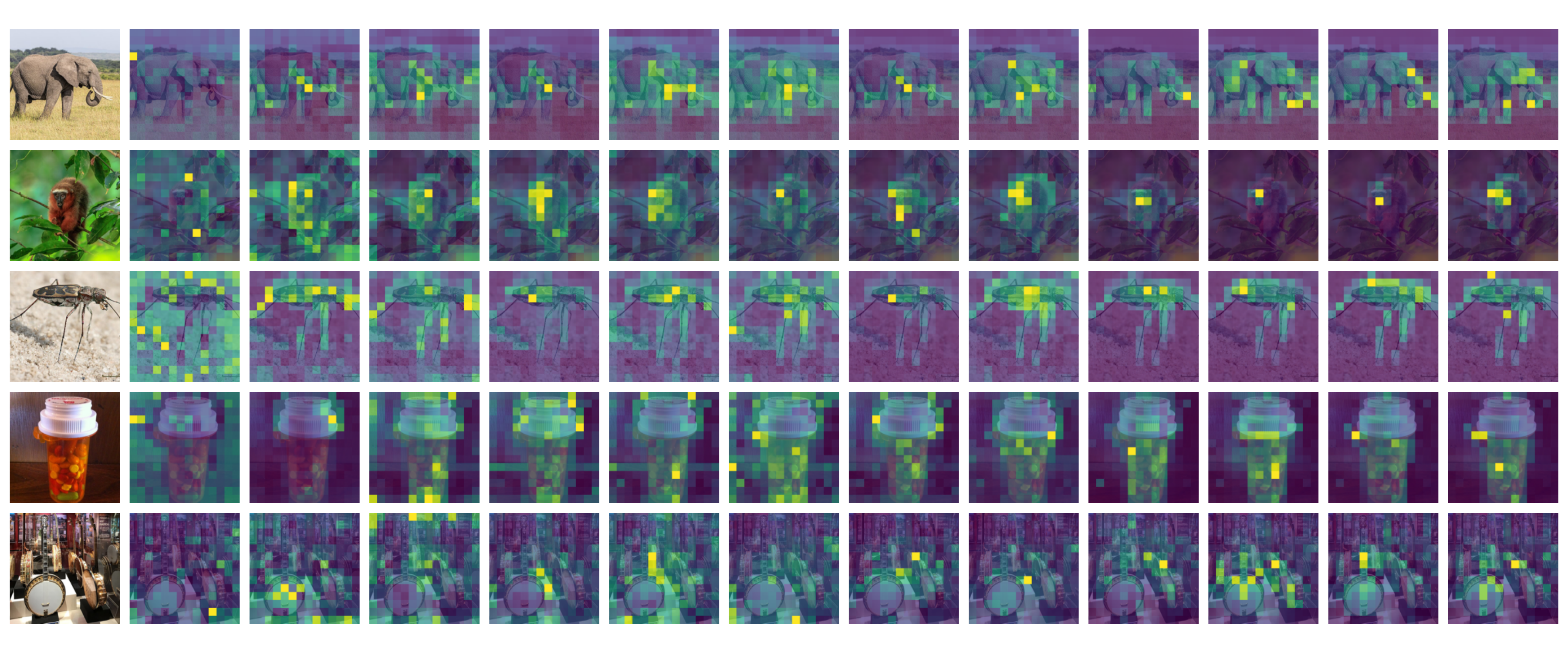} % , height=220

    \caption{Attention maps across layers, from first to twelfth layer (left to right respectively). All images are from Wikimedia Commons under CC BY or CC BY-SA licenses \cite{foubister_4,parentingpatch_pill,karim_lophyra,jacqke_vega,hobbyfotowiki_african,}.}
    \label{figS:Examples attmaps through layers}

\end{figure}

\begin{figure}[ht]
    \centering
    \makebox[\textwidth][c]{%
        \begin{subfigure}{\textwidth}
            \includegraphics[width=\linewidth]{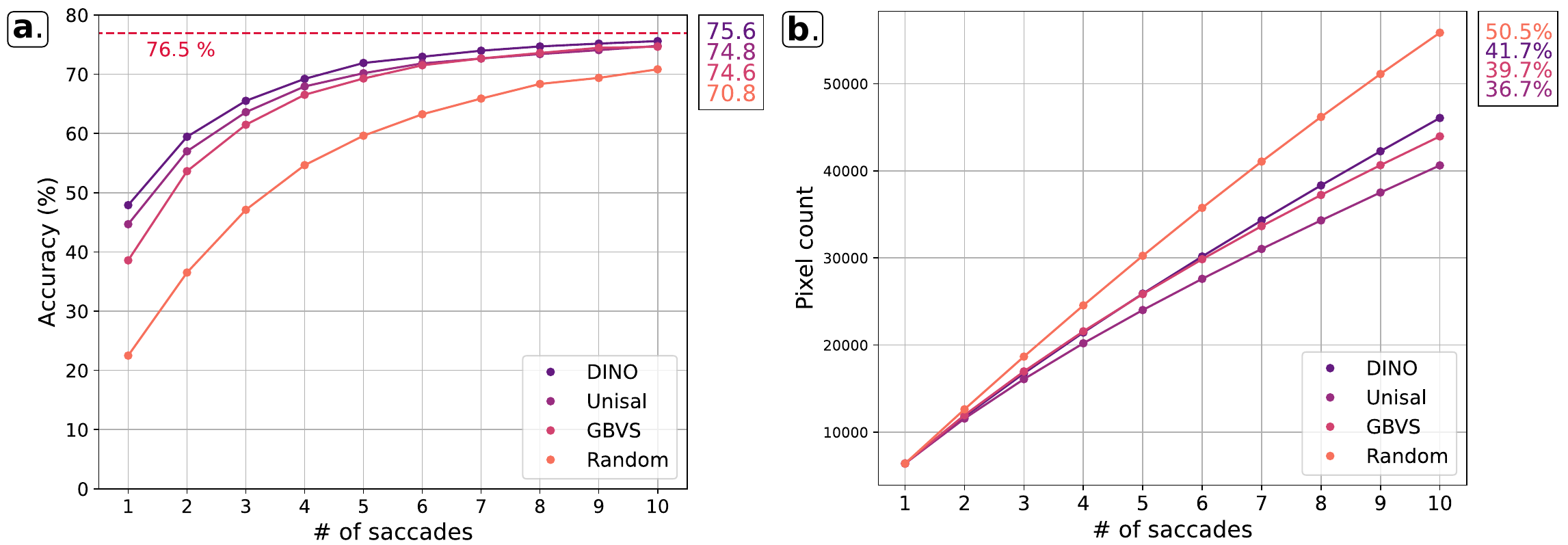}
        \end{subfigure}

    }

    \caption{Accuracy score over multiple saccades driven by attention maps from different models, for fovea of size $5\times{}5$ tokens (\textbf{a.}) and the corresponding percentage of the image given to the model throughout saccades (\textbf{b.})}
    \label{figS:Accuracy}
\end{figure}

\begin{figure}[ht]

    \begin{subfigure}{\textwidth}
    \includegraphics[width=\linewidth]{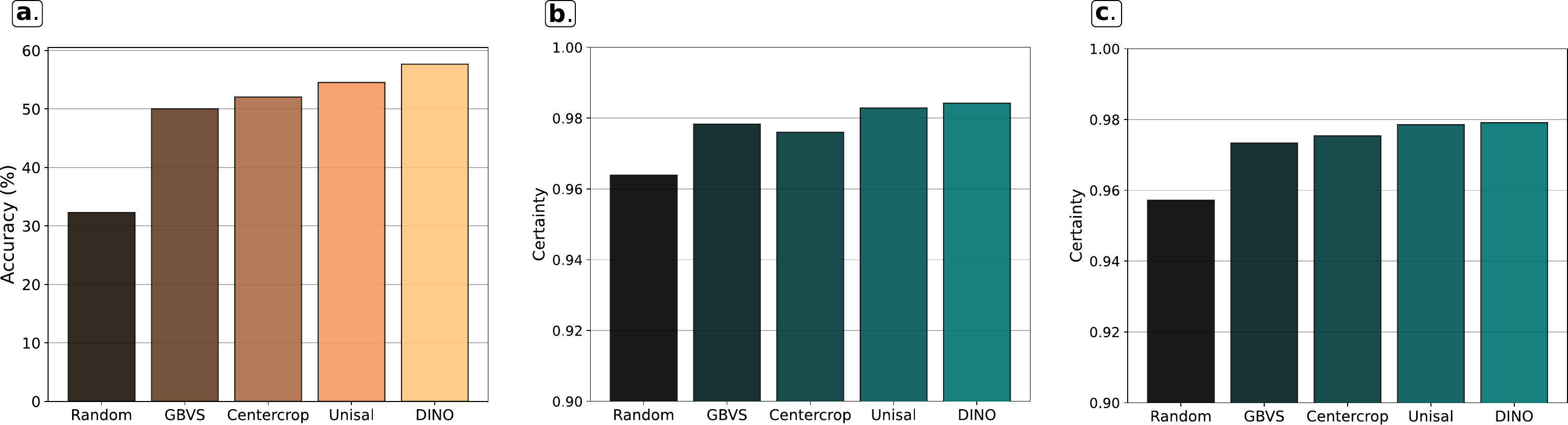} 

    \end{subfigure}

\caption{Accuracy in CNN classification for $80\times{}80$ pixels crops (~5.8\% of the original image) (\textbf{a.}). Mean certainty in CNN classification across condition for $48\times{}48$ pixels crops (\textbf{b.}) and $80\times{}80$ pixels crops (\textbf{c.}).}
\label{figS:Accuracy Certainty CNN across models Supp}
\end{figure}

\begin{figure}[ht]
    \centering
    \begin{subfigure}[b]{\linewidth}
        \centering
        \includegraphics[width=\linewidth]{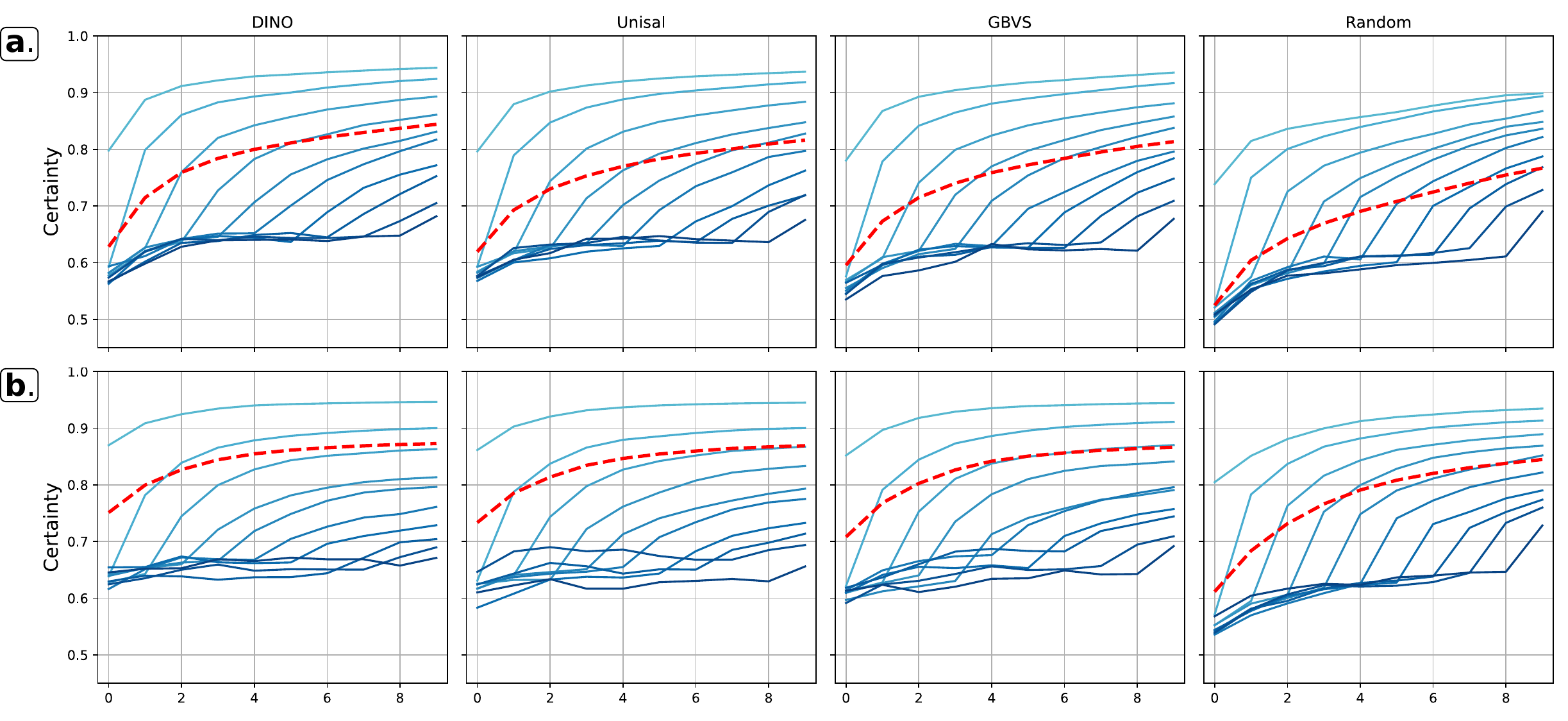}
    \end{subfigure}
    
    \caption{Evolution of the model’s classification certainty across saccades: Curves represent the mean certainty across the different saliency models, and the random fixation condition. Colors indicate sets of images correctly classified at a given saccade (dark to light blue: first to tenth saccade). The red dotted line shows the mean certainty of all images, across saccade. Colored bars in the rectangles on the right of each plot shows mean certainty when the full image is given to the model, for each set. For fovea of size \(3 \times 3\) tokens (\textbf{a.}) and \(5 \times 5\) token (\textbf{b.})}
    \label{fig:certainty_models}
\end{figure}

\begin{figure}[ht]

    \begin{subfigure}{\textwidth}
    \includegraphics[width=\linewidth]{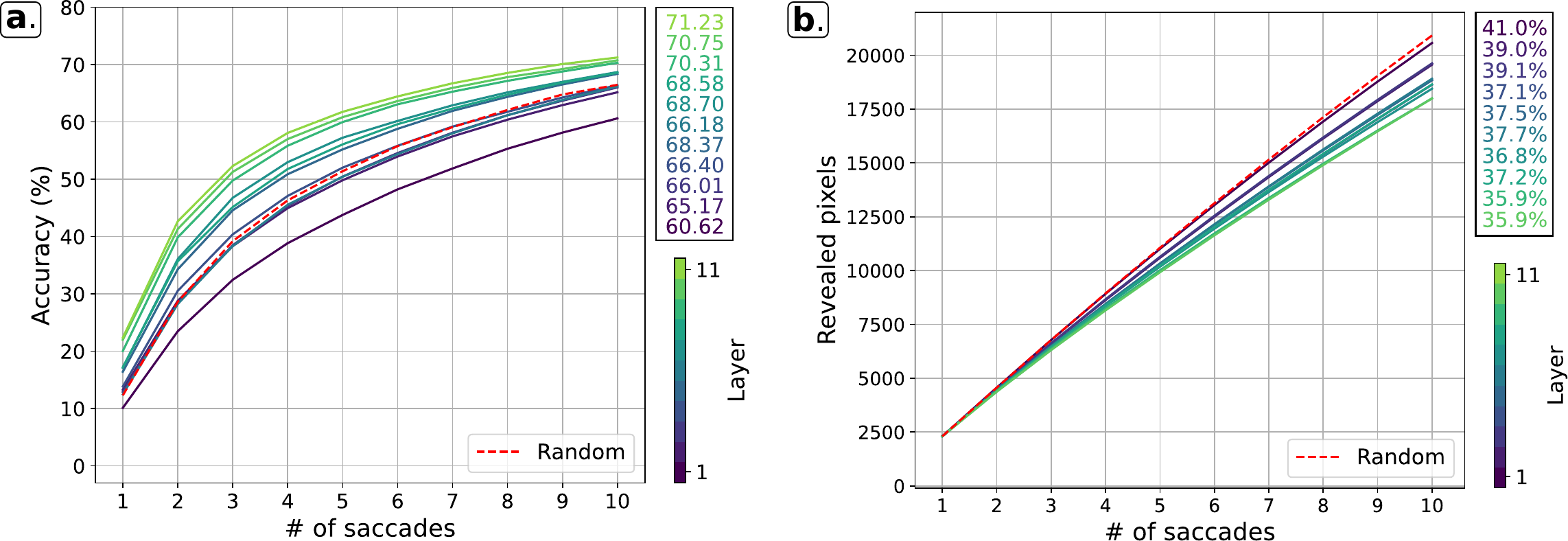} 
    \end{subfigure}

\caption{Accuracy score across saccades for attention maps of low-resolution $128\times{}128$ pixel input, taken at different layers (\textbf{a.}), and the corresponding percentage of the image given to the model throughout saccades (\textbf{b.}).}
\label{figS:Accuracy across layers low res}
\end{figure}

\

\clearpage

\printbibliography

\end{document}